\UseRawInputEncoding
\documentclass[letterpaper, 10 pt, conference]{ieeeconf}  % Comment this line out if you need a4paper

\IEEEoverridecommandlockouts                              % This command is only needed if 
\usepackage{graphicx} % for pdf, bitmapped graphics files
\usepackage{cite}
\usepackage{amsmath}

\title{\LARGE \bf
PerchMobi\textsuperscript{3}: A Multi-Modal Robot with Power-Reuse Quad-Fan Mechanism for Air-Ground-Wall Locomotion
}
 \author{Yikai Chen$^{1,2,3}$, Zhi Zheng$^{1,2,3}$, Jin Wang$^\dag$$^{1,2,3}$, Bingye He$^{1,2,3}$, \\Xiangyu Xu$^{1,2,3}$, Jialu Zhang$^{1,2,3}$, Huan Yu$^{4}$ and Guodong Lu$^{1,2,3}$
 \thanks{This work was supported in part by the National Natural Science Foundation of China under Grant 52475033, in part by the ``Pioneer" and ``Leading Goose" R\&D Program of Zhejiang under Grant 2024C01170, and in part by the Robotics Institute of Zhejiang University under Grant K12107 and Grant K11805. \textit{(Yikai Chen and Zhi Zheng contributed equally to this work.)}}
 \thanks{
 	$^{1}$The State Key Laboratory of Fluid Power and Mechatronic Systems, School of Mechanical Engineering, Zhejiang University, Hangzhou 310058, China. $^{2}$Zhejiang Key Laboratory of Industrial Big Data and Robot Intelligent Systems, Zhejiang University, Hangzhou 310058, China. $^{3}$Robotics Research Center of Yuyao City, Ningbo 315400, China. $^{4}$College of Control Science and Engineering, Zhejiang University, Hangzhou 310027, China.
 }
 \thanks{Email: {\tt\small dwjcom@zju.edu.cn}}
 \thanks{\dag Corresponding author: Jin Wang.}
 }

\begin{document}

\maketitle
\thispagestyle{empty}
\pagestyle{empty}

\begin{abstract}
Achieving seamless integration of aerial flight, ground
driving, and wall climbing within a single robotic platform
remains a major challenge, as existing designs often rely
on additional adhesion actuators that increase complexity,
reduce efficiency, and compromise reliability. To address
these limitations, we present PerchMobi\textsuperscript{3}, a quad-fan,
negative-pressure, air-ground-wall robot that implements a
propulsion-adhesion power-reuse mechanism. By repurposing
four ducted fans to simultaneously provide aerial thrust
and negative-pressure adhesion, and integrating them with
four actively driven wheels, PerchMobi\textsuperscript{3} eliminates dedicated pumps while maintaining a lightweight and compact design. To the best of our knowledge, this is the first quad-fan prototype to demonstrate functional power reuse for multi-modal locomotion. A modeling and control
framework enables coordinated operation across ground,
wall, and aerial domains with fan-assisted transitions. The
feasibility of the design is validated through a
comprehensive set of experiments covering ground driving,
payload-assisted wall climbing, aerial flight, and
cross-mode transitions, demonstrating robust adaptability
across locomotion scenarios. These results highlight the
potential of PerchMobi\textsuperscript{3} as a novel design paradigm for multi-modal robotic mobility, paving the way for future extensions toward autonomous and application-oriented
deployment.
\end{abstract}

\section{Introduction}
In recent years, ground-air-wall robots have achieved remarkable progress in academic, industrial, and military applications \cite{david2021design, dias2023bogiecopter, liu2023hitchhiker}. These robots integrate multiple locomotion capabilities—such as aerial flight, ground driving, and wall climbing—combining the long-endurance advantage of wall-climbing robots with the cross-domain mobility of aerial platforms. However, the efficient integration of multiple movement modes into a single robotic system still poses significant design challenges, and further research is urgently needed in areas such as structural optimization, control strategies, and energy management. Motivated by these challenges, this work proposes PerchMobi\textsuperscript{3}, a novel quad-fan air-ground-wall robot that implements a propulsion-adhesion power reuse strategy and develops control schemes for coordinated multi-modal locomotion.

\begin{figure}[t]
	\begin{center}
		\includegraphics[width=1.0\columnwidth]{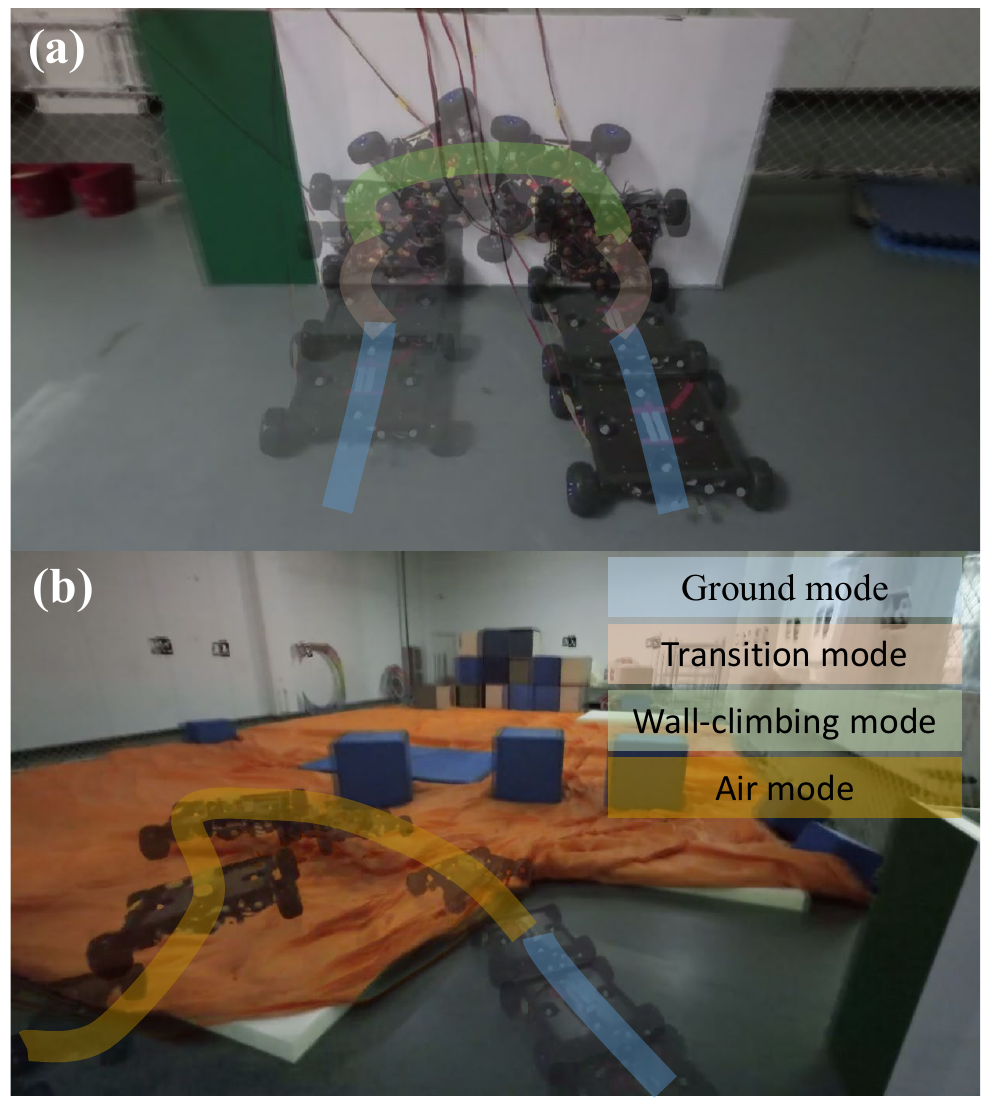}
	\end{center}
	\vspace{-0.5cm}
	\caption{
		\label{fig:abstract}
        Experimental demonstrations of multi-modal locomotion. (a) Ground-to-wall transition with lateral motion and return to the ground, illustrating traversal through a narrow gap. (b) Ground-to-flight transition, illustrating obstacle negotiation by vertical takeoff and flight.
	}
	\vspace{-0.2cm}
\end{figure}

To address the aforementioned challenges, researchers have proposed various designs for ground-air-wall robots, extensively integrating mechanisms such as flight, crawling, and adhesion. Among these, negative-pressure adhesion technology has emerged as a common choice due to its relative maturity. Existing wall-climbing robots with active adhesion mechanisms \cite{tao2023design, ang2022drone, chapdelaine2024simulation, hirano2024autonomous, sahu2024design} often rely on devices such as electric pumps \cite{tsukagoshi2015aerial, liu2020adaptive} or inclined structures \cite{dautzenberg2023perching, lee2023minimally, lee2024autonomous}. However, such solutions complicate the control system, reduce reliability, and increase both manufacturing and maintenance costs. Conversely, passive adhesion mechanisms—including suction cups \cite{hsiao2022novel, nishio2021fixed}, clamping mechanisms \cite{lee2024rapid, lan2024aerial, lynch2024powerline}, and self-sealing structures \cite{qin2022design, wopereis2016mechanism}—can achieve stable adhesion, but typically require dedicated adhesion modules. This significantly increases structural complexity and weight, while also degrading the overall control accuracy of the system. Therefore, there is an urgent need for a more efficient integration of propulsion, structure, and control to enable lightweight, reliable, and versatile multi-modal robots.

In contrast, reusing specific actuators from the flight configuration—such as ducted fans—for wall locomotion offers the potential to reduce the need for additional mechanisms, thereby enabling a more compact and efficient cross-domain robot design. Analysis of typical negative-pressure wall-climbing robots reveals that ducted fans can serve dual functions: providing propulsion in a quadcopter configuration and generating negative-pressure adhesion forces. This suggests that functional reuse of power units can lead to more efficient cross-domain mobility mechanisms for ground-air-wall robots. Building on this concept, this study develops a composite system that integrates ducted fans with negative-pressure cavity plates, enabling flight thrust in aerial mode, adhesion in wall-climbing mode, and power-assisted transitions between ground and wall locomotion. The detailed design and implementation are presented in Section~\ref{sec:Design}.

\begin{figure}[t]
	\begin{center}
		\includegraphics[width=1.0\columnwidth]{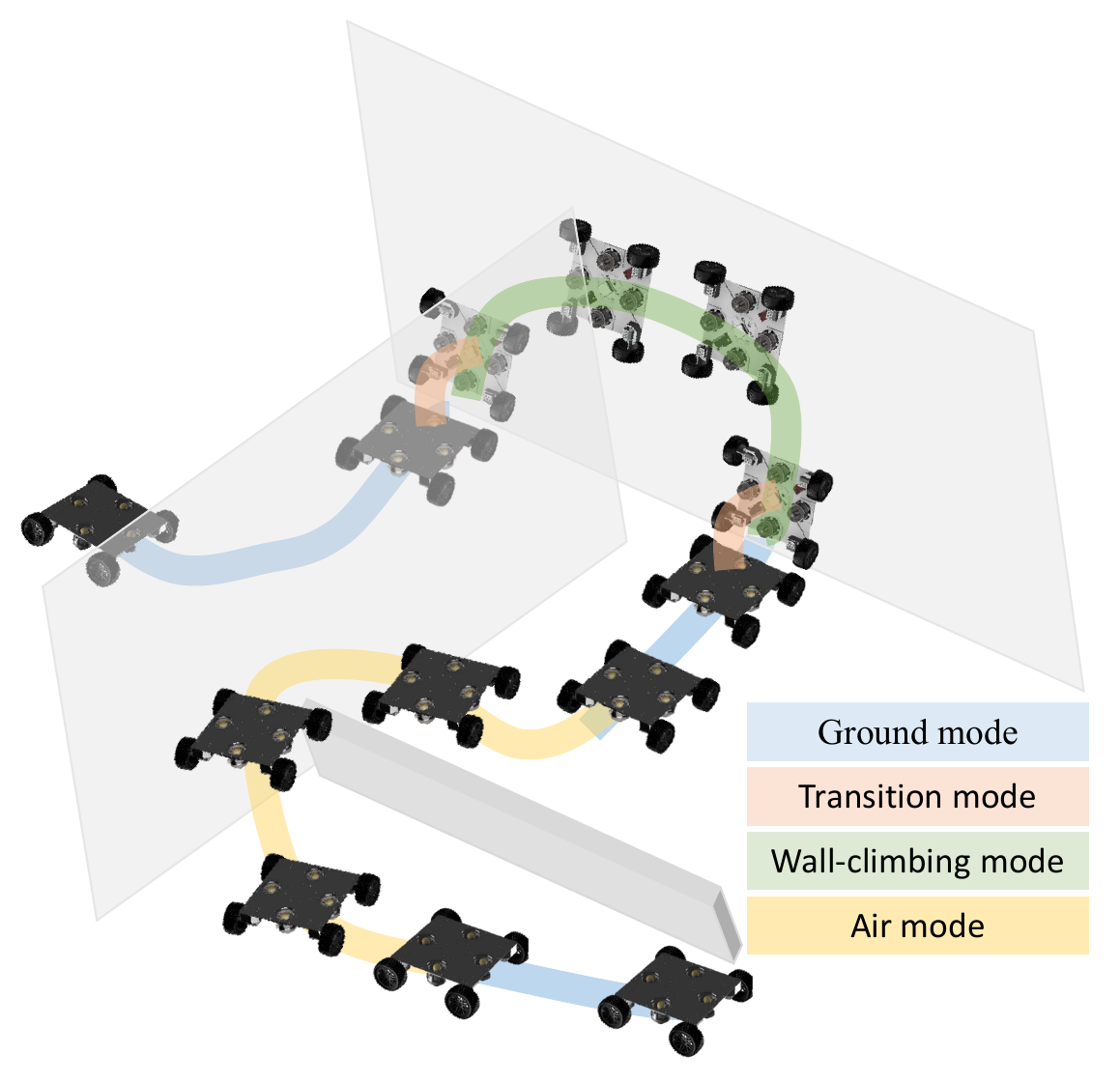}
	\end{center}
        \vspace{-0.5cm}
	\caption{
		\label{fig:introduction}
		Ground-to-wall transition with lateral motion and return to the ground, demonstrating traversal through a narrow gap, and ground-to-flight transition, demonstrating obstacle negotiation by vertical takeoff and flight. These experiments correspond to the experimental scenarios illustrated in Fig.~\ref{fig:abstract}.
	}
        \vspace{-0.2cm}
\end{figure}

In terms of wall adhesion stability, early studies primarily employed active adhesion mechanisms such as electric suction cups or motor-driven pumps, which provide strong adhesion forces and vacuum extraction capabilities. Among these, electric suction cups have been widely adopted due to their relatively simple control schemes \cite{tao2023design, ang2022drone, chapdelaine2024simulation, hirano2024autonomous, sahu2024design, tsukagoshi2015aerial, liu2020adaptive, hsiao2022novel, nishio2021fixed, qin2022design, wopereis2016mechanism}. However, compared with traditional wall-climbing robots of the same weight and size, vacuum-based schemes using motor pumps or ducted fans typically achieve higher energy efficiency \cite{zhang2022novel} and better obstacle-crossing performance on rough wall surfaces \cite{siyuan2021research}. In recent years, negative-pressure wall-climbing robots integrating motor pumps \cite{wu2018research, zhang2017study, brusell2017novel, liang2013design, wang2013development} or ducted fans \cite{sekhar2014duct, inoue2018inspection} have demonstrated notable progress. Nevertheless, existing ground-air-wall robots have not effectively addressed the increased energy consumption and control complexity introduced by additional adhesion mechanisms, which significantly limit their performance in wall-climbing modes. In this context, the design approach of reusing four ducted fans for both propulsion and adhesion presents a promising solution, as it can simultaneously enable stable aerial flight and reliable wall adhesion, highlighting its potential for cross-domain motion integration. Inspired by this concept, this work proposes a composite robotic design that integrates four ducted fans with four actively driven wheels, thereby achieving multi-modal locomotion—including stable aerial flight, robust wall climbing with obstacle negotiation, and ground driving—without relying on dedicated adhesion modules.

Based on the above analysis, this paper proposes a novel design scheme for a ground-air-wall multi-modal robot. The scheme combines the aerial maneuverability of quadrotor platforms with the wall stability of negative-pressure climbing robots, while incorporating four actively driven wheels to enable ground locomotion and obstacle negotiation. The core innovation lies in the power reuse strategy, whereby four cross-shaped ducted fans simultaneously provide flight thrust in aerial mode and generate negative-pressure adhesion in wall-climbing mode. Building upon this strategy, the PerchMobi\textsuperscript{3} prototype achieves efficient integration of aerial, ground, and wall mobility through the coordinated operation of four ducted fans and four steering servos. Compared with conventional negative-pressure wall-climbing robots, the proposed system ensures stable wall adhesion without relying on dedicated suction motors, while maintaining the ability to perform controllable flight. With its durability, motion stability, obstacle-crossing capability, and cross-domain maneuverability, Such multi-modal mobility is inspired by representative scenarios such as building façades and wind turbine structures, which require both aerial access and stable wall locomotion.

To fully exploit the multi-modal locomotion potential of PerchMobi\textsuperscript{3}, we developed a dynamic modeling and control framework that integrates the quadrotor dynamic model with the four-wheel self-balancing robot model, governing aerial, ground, and wall modes. A full-scale prototype was constructed, and its feasibility was validated through experiments. The results confirm that the robot achieves high motion stability and reliable obstacle-crossing capability in wall-climbing mode, demonstrating the adaptability and execution potential of PerchMobi\textsuperscript{3} in realistic and complex environments.

The main contributions of this work are summarized as follows:
\begin{itemize}
	\item We introduce PerchMobi\textsuperscript{3}, a novel multi-modal robot that integrates a \emph{power-reuse quad-fan mechanism}, in which four ducted fans are repurposed to simultaneously provide aerial thrust and generate negative-pressure adhesion. This design eliminates the need for dedicated suction pumps while maintaining compactness and lightweight structure.
	
	\item We establish dynamic modeling and control frameworks that enable coordinated operation across ground, wall, and aerial domains, including fan-assisted transitions. The feasibility of these schemes is validated through a full-scale prototype.
	
	\item We conduct comprehensive experimental demonstrations, including ground driving, payload-assisted wall climbing, aerial flight, and cross-mode transitions, thereby confirming the effectiveness, robustness, and adaptability of the proposed design in diverse locomotion scenarios.
\end{itemize}

\section{Design of PerchMobi\textsuperscript{3}}\label{sec:Design}
\subsection{Mechanical Structure Design}
\begin{figure}[t]
	\begin{center}		
    \includegraphics[width=1.0\columnwidth]{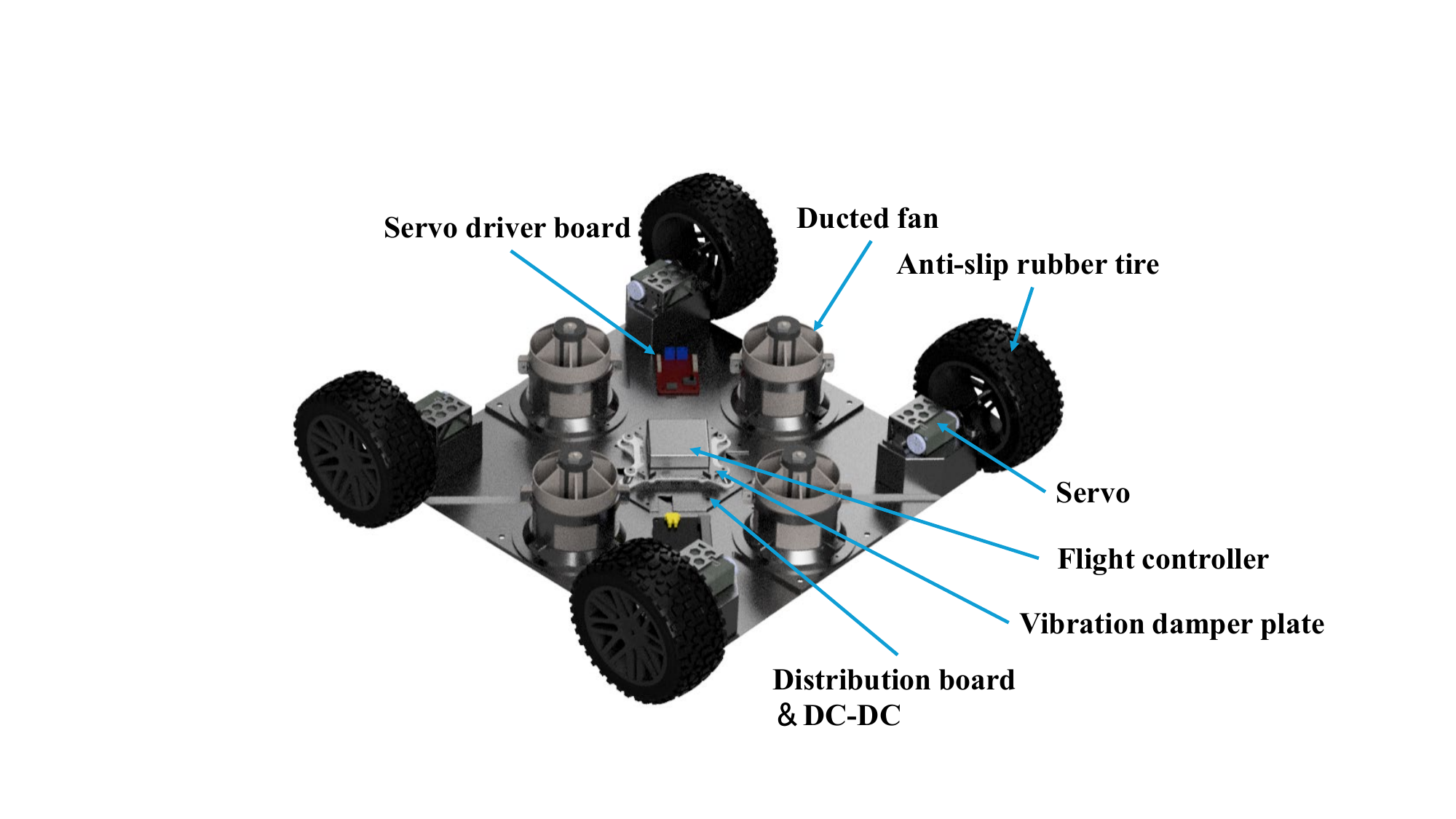}
	\end{center}
	\vspace{-0.5cm}
	\caption{
		\label{fig:Mechanical}
		Mechanical design of PerchMobi\textsuperscript{3}, highlighting the integrated carbon-fiber frame, ducted fans, wheel drives, sealing elements, and protective components.
	}
	\vspace{-0.2cm}
\end{figure}

The mechanical design of PerchMobi\textsuperscript{3}, emphasizing multi-modal adaptability, lightweight yet high rigidity, and coordinated adhesion-mobility, is illustrated in Fig.\ref{fig:Mechanical}. The prototype is built around an integrated carbon-fiber frame and incorporates ducted fans, wheel drives, sealing elements, and protective components. Its feasibility has been validated through operation in four distinct modes: ground locomotion, wall climbing, aerial flight, and mode transition. The overall system weight of the prototype is $3.2 \mathrm{kg}$.

The main bearing frame adopts a high-modulus carbon fiber plate as the core structure. The plate is functionally divided into two surfaces:
\begin{itemize}
\item Surface A (electronics components side) is milled for weight reduction, with the non-load-bearing regions thinned to $1 \mathrm{mm}$, and provides mounting holes for ducted fans (4 $\times$ $70 \mathrm{mm}$), servos (4 sets), and power distribution board supports.
\item Surface B (negative pressure adsorption surface) corresponds to the non-milled side of the carbon-fiber plate and includes a sealing sponge groove along the outer edge. Structural optimization ensures effective sealing during adsorption.
\end{itemize}

\begin{figure}[t]
	\begin{center}
		\includegraphics[width=1.0\columnwidth]{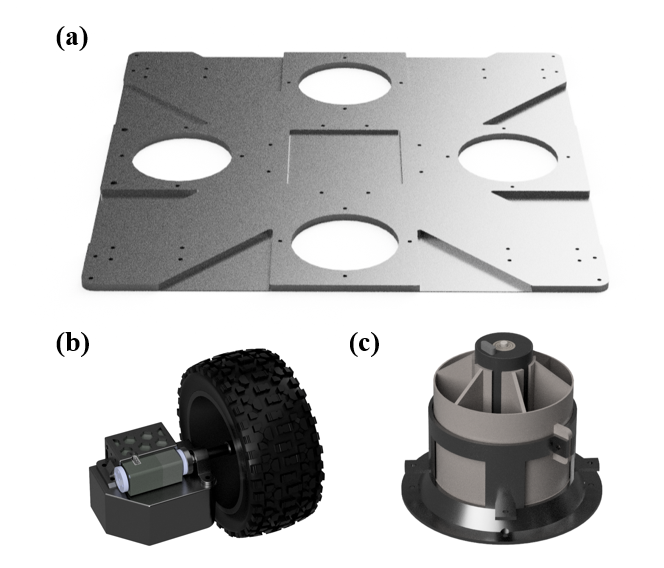}
	\end{center}
	\vspace{-0.5cm}
	\caption{
		\label{fig:Compoments}
    Components of PerchMobi\textsuperscript{3}. (a) Cross-shaped carbon plate. (b) Wheel drive assembly. (c) Ducted fan assembly.
	}
	\vspace{-0.2cm}
\end{figure}

As illustrated in Fig.\ref{fig:Compoments}, the ducted fans, servos, and central carbon plate together form a cross-shaped configuration, which enhances the overall structural strength of the UAV while supporting lightweight design.

The ducted fan assembly constitutes the primary power unit for both negative-pressure adhesion and aerial flight. Four brushless ducted fans ($\phi 70\mathrm{mm}$, model QF3027-2200KV, $209 \mathrm{g}$ each, maximum thrust $3300 \mathrm{g}$) are mounted directly on surface~A of the carbon-fiber plate. Adhesive sealing is applied along the joint interface to enhance fixation and prevent air leakage.

The wheel drive assembly provides locomotion in both ground and wall-climbing modes. It consists of four digital servos (Feetech SYS3025BL, $89 \mathrm{g}$, torque $40 \mathrm{kg\cdot cm}$), four elastic rubber tires ($\phi 130 \mathrm{mm}$, $130 \mathrm{g}$ each), and aluminum couplings. The servos are mounted at the four corners of surface~A and connected to the tires via couplings. On the ground, the tires maintain a $0.5 \mathrm{mm}$ clearance between surface~B and the ground, preventing sponge strip abrasion and reducing slippage. During wall transitions, the elasticity of the tires provides buffering, thereby protecting the carbon-fiber plate from impact damage.

\subsection{Electrical System Design}
\begin{figure}[t]
	\begin{center}
		\includegraphics[width=1.0\columnwidth]{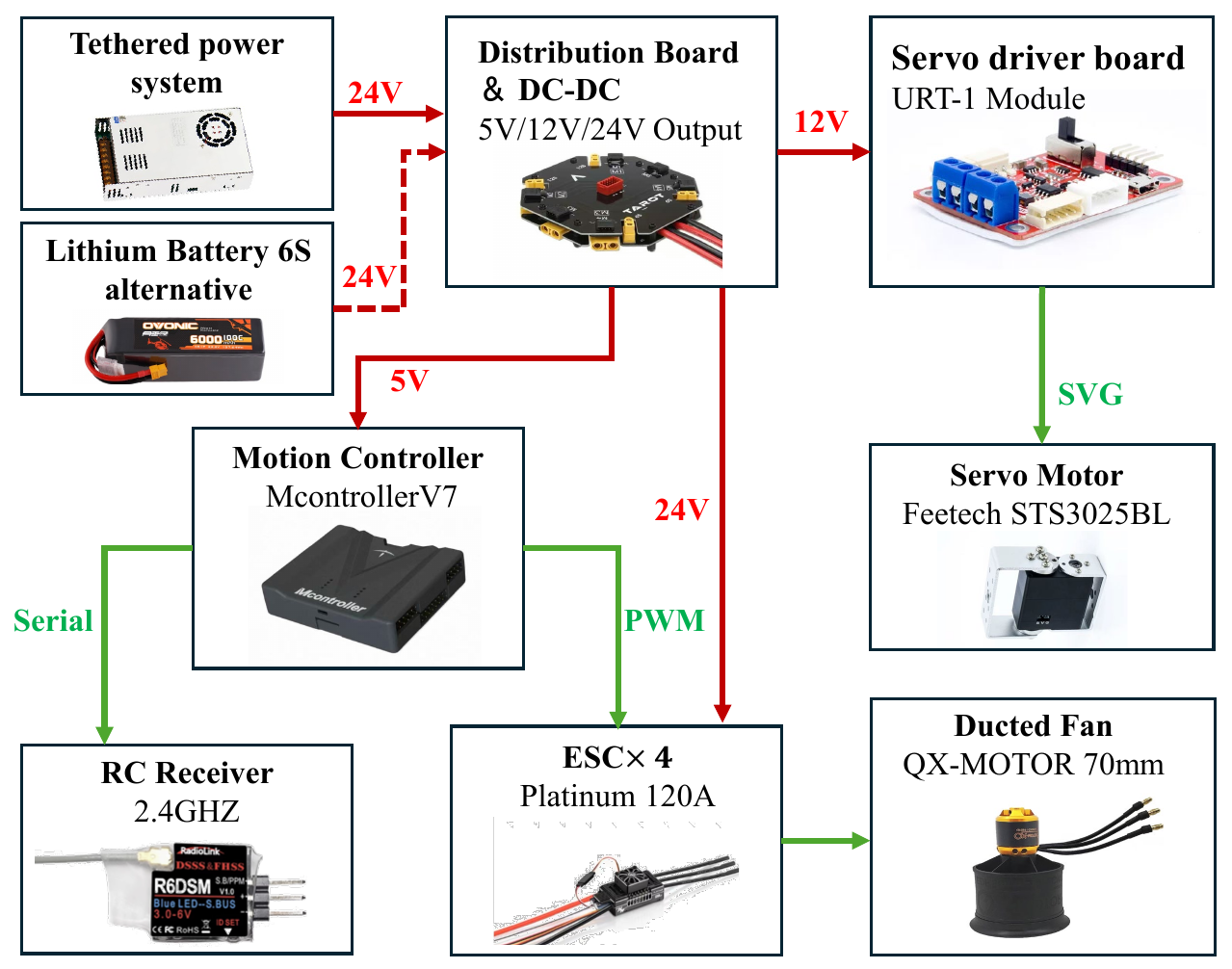}
	\end{center}
	\vspace{-0.5cm}
	\caption{
		\label{fig:Eletronic}
		PerchMobi\textsuperscript{3} electrical system architecture, including power supply, flight controller, ESCs, and servo driver.
	}
	\vspace{-0.2cm}
\end{figure}

As the core enabler of system functionality, the electrical system is designed according to the principles of “stable power supply, precise control, and low-interference transmission.” With a modular layout and coordinated control strategy, it ensures reliable power delivery and real-time signal processing to support multi-mode operation. Installation tests verify both system stability and dynamic response. The electrical architecture of PerchMobi\textsuperscript{3} is illustrated in Fig.~\ref{fig:Eletronic}.

The power system primarily adopts a tethered power supply scheme, while a 6S high-energy-density lithium-polymer battery can also be used as an alternative source. Power is distributed through a central power distribution board located at the center of surface~A. The board connects directly to the power source and delivers power to the flight controller ($5 \mathrm{V}/2 \mathrm{A}$), brushless ESCs (without voltage step-down), and the servo driver board ($12 \mathrm{V}/5 \mathrm{A}$), thereby ensuring voltage compatibility and avoiding power conflicts.

For multimodal motion control, we employ a microcontroller featuring an STM32F427 ARM processor with an integrated IMU chip (ICM20608). PerchMobi\textsuperscript{3} is operated via a remote controller, while Kalman filtering is applied to update orientation estimates at $400 \mathrm{Hz}$.

The execution drive system consists of four $120 \mathrm{A}$ brushless ESCs and a servo driver board. Each ESC independently controls one ducted fan and, upon receiving PWM signals, enables continuous adjustment of thrust and negative pressure. The servo driver board performs signal conversion from the controller for the four wheel servos, supporting both speed control and position control.

\subsection{Modeling and Control Design}

The robotic system operates in four distinct modes: ground inspection, wall climbing, ground-to-wall transition, and aerial flight. Each mode is governed by a dedicated kinematic model and control strategy, formulated with respect to appropriate coordinate frames and actuation mechanisms.

\subsubsection{Ground Mode}
In this mode, the robot performs wheeled locomotion on flat terrain. Let the world frame $O_{xyz}^G$ be fixed with $X$ forward, $Y$ lateral, and $Z$ upward. The body-fixed frame $O_{x'y'z'}^G$ is centered at the carbon plate, with $x'$ aligned with the longitudinal axis. Assuming a differential drive configuration, the linear and angular velocities are:
\begin{equation}
    \label{eq:ground_vel}
    \begin{aligned}
    U &= \frac{r}{2} (\omega_L + \omega_R), \\
    \omega &= \frac{r}{L} (\omega_R - \omega_L),
    \end{aligned}
\end{equation}
where $r$ is the wheel radius, $L$ the track width, and $\omega_{L/R}$ the left/right wheel angular speeds. The pose evolution is governed by:
\begin{equation}
    \label{eq:ground_kin}
    \dot{\theta} = \omega, \quad
    \dot{X} = U \cos\theta, \quad
    \dot{Y} = U \sin\theta.
\end{equation}

\subsubsection{Wall Climbing Mode}
During vertical wall traversal, negative pressure adhesion is activated. The wall-fixed frame $O_{xyz}^W$ has $X$ horizontal, $Y$ vertical upward, and $Z$ outward normal to the wall. The body frame $O_{x'y'z'}^W$ aligns with the surface (with $z'$ toward the wall). Horizontal and vertical wheel pairs provide motion:
\begin{equation}
    \label{eq:wall_vel}
    \begin{aligned}
    U' &= \frac{r}{2}(\omega_{L1} + \omega_{R1}), \\
    V' &= \frac{r}{2}(\omega_{L2} + \omega_{R2}),
    \end{aligned}
\end{equation}
and steering is achieved via differential drive:
\begin{equation}
    \label{eq:wall_omega}
    \omega' = \frac{r}{2L} \left[ (\omega_{R1} - \omega_{L1}) + (\omega_{R2} - \omega_{L2}) \right].
\end{equation}

Roll and pitch deviations ($\Delta\alpha, \Delta\beta$) due to surface irregularities are compensated by adjusting fan speeds:
\begin{equation}
    \label{eq:wall_compensation}
    \Delta \boldsymbol{\omega}_{\text{fan}} = K_c (\Delta\alpha + \Delta\beta).
\end{equation}

\subsubsection{Flight Mode}
Aerial motion is driven by four ducted fans located at the corners of surface $a$. In the body frame, each fan generates lift:
\begin{equation}
    \label{eq:lift_model}
    F_i = k_f \omega_i^2,
\end{equation}
the net vertical force and body moments are:
\begin{equation}
    \label{eq:total_force}
    F_z = \sum_{i=1}^4 F_i - mg,
\end{equation}
\begin{equation}
    \label{eq:body_moments}
    \boldsymbol{M} = 
    \begin{bmatrix}
    M_x \\ M_y \\ M_z
    \end{bmatrix}
    =
    \begin{bmatrix}
    \dfrac{d}{2}(F_2 + F_4 - F_1 - F_3) \\[2ex]
    \dfrac{d}{2}(F_1 + F_2 - F_3 - F_4) \\[2ex]
    k_m (F_2 + F_3 - F_1 - F_4)
    \end{bmatrix},
\end{equation}
where $d$ is the half-diagonal distance, and $mg$ is the gravitational force.

The translational and rotational kinematics are:
\begin{subequations}
    \label{eq:flight_kinematics}
    \begin{align}
    \dot{X} &= U \cos\gamma - V \sin\gamma, \\
    \dot{Y} &= U \sin\gamma + V \cos\gamma, \\
    \dot{Z} &= W, \\
    \dot{\alpha} &= p, \quad \dot{\beta} = q, \quad \dot{\gamma} = r,
    \end{align}
\end{subequations}
where $(U,V,W)$ and $(p,q,r)$ are body-fixed linear and angular velocities, and $\gamma$ is the yaw angle.

\subsubsection{Mode Transition: Ground-to-Wall}
The transition is executed in four stages: wall approach, tire buffering, airframe flipping, and adsorption engagement.

\begin{itemize}
    \item \textbf{Airframe Flipping:} Rotation about wall base $(X_0,0,0)$ with angular dynamics:
    \begin{equation}
        \label{eq:flip_dynamics}
        \dot{\theta}_{\text{flip}} = \frac{M_{\text{fan}}}{I_z}.
    \end{equation}
    \item \textbf{Adsorption Engagement:} Negative pressure builds as:
    \begin{equation}
        \label{eq:pneg_dynamics}
        \dot{P}_{\text{neg}} = k_p (\omega_{\text{fan}} - \omega_{\text{set}}),
    \end{equation}
    where $k_p$ is the pressure gain coefficient, and $\omega_{\text{set}}$ is the target fan speed.
\end{itemize}

\section{Experiments and Results}

\subsection{Single-Scenario Locomotion}
\begin{figure}[t]
	\begin{center}
		\includegraphics[width=1.0\columnwidth]{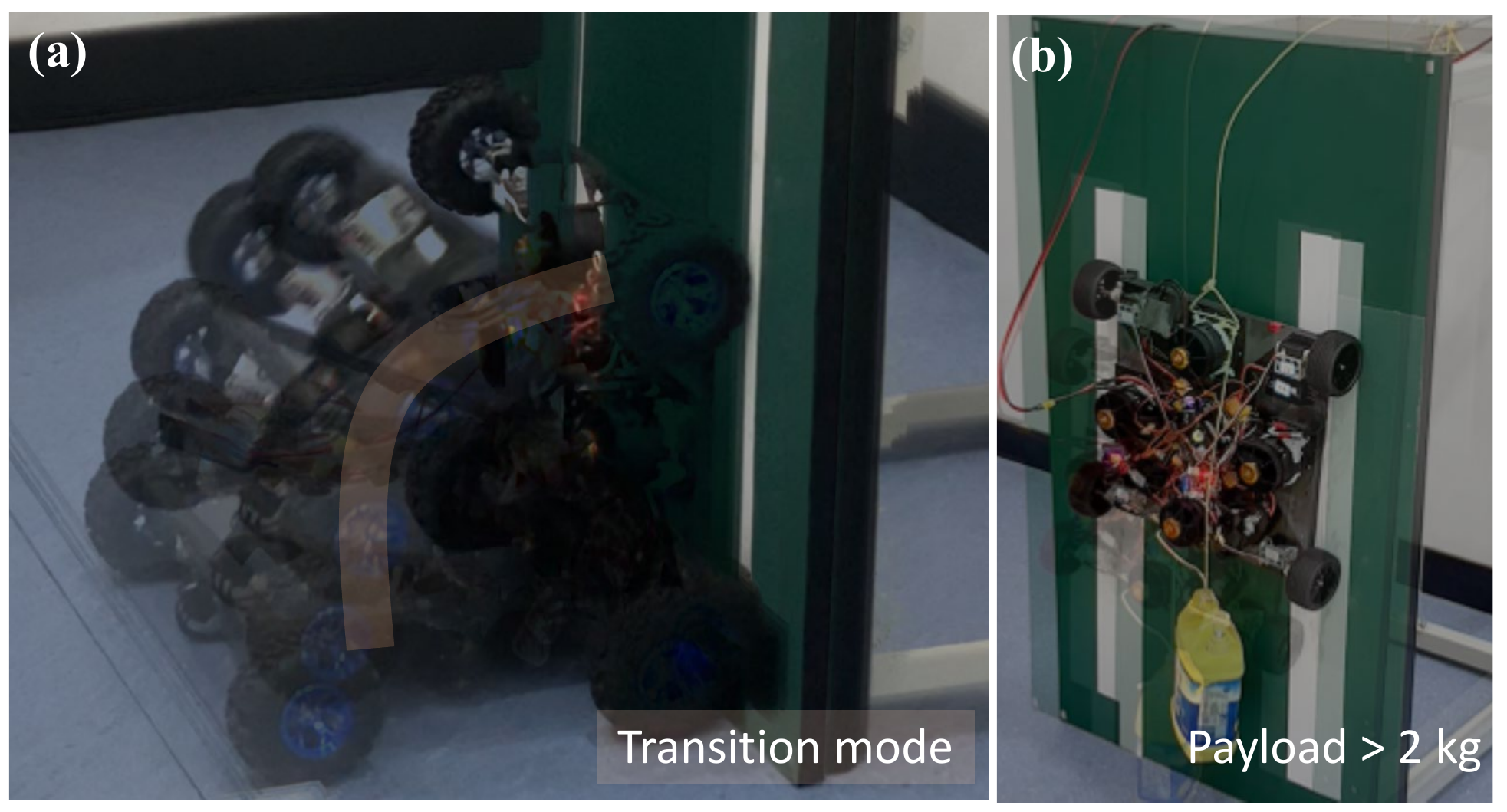}
	\end{center}
        \vspace{-0.5cm}
	\caption{
		\label{fig:transition}
		(a) Ground-to-wall transition experiment. (b) Wall-climbing experiment with payloads, demonstrating stable upward locomotion with loads greater than $2 \mathrm{kg}$.
	}
        \vspace{-0.2cm}
\end{figure}

We constructed the PerchMobi\textsuperscript{3} prototype platform, which is capable of efficient locomotion on the ground, in the air, and along vertical walls. The feasibility of these functions has been validated through experimental demonstrations in ground driving, aerial flight, and wall-climbing scenarios. Wall-climbing experiments with payloads, as shown in Fig.~\ref{fig:transition}(b), demonstrate stable motion with loads exceeding $2 \mathrm{kg}$, while aerial flight experiments confirm the feasibility of controlled flight. All experimental images are presented as multi-frame composites.

\subsection{Transitional Locomotion}

In addition to single-mode locomotion, the robot achieves smooth transitions from the ground to the wall and from the ground to the air by regulating the rotational speed of the ducted fans. To verify the feasibility of the proposed control algorithm, experiments were conducted demonstrating stable transitions from a horizontal ground to a vertical wall and from a horizontal ground to the air. A supplementary video provides a comprehensive demonstration of these transitions, and multi-frame composite images are shown in Fig.~\ref{fig:transition}(a).

The ground-to-wall transition consists of two stages: flipping and reinforced wall adsorption. In the flipping stage, three outer ducted fans are activated to generate asymmetric torque, causing the robot to lift and rotate smoothly around its bottom edge. After the start command is issued, the three outer fans accelerate to their initial thrust setpoints with a preset ramp rate. Among them, the fan farthest from the wall provides slightly higher thrust to produce a pitching moment that lifts the robot’s front end, while the inner fan closest to the wall remains inactive during this stage. Driven by this torque, the airframe rotates backward about the $Y$-axis (pitch axis), and the pitch angle gradually increases.

In the reinforced adsorption stage, the ducted fan speeds are carefully regulated to guide the robot into stable wall attachment. The controller gradually reduces the thrust of the three outer fans, allowing the body to continue rotating under gravity with lower speed and acceleration. Based on attitude and distance feedback, the fan thrust is dynamically adjusted to suppress excessive angular velocity. As soon as the robot reaches a vertical attitude, the sealed chamber contacts the wall and negative pressure rapidly builds up. Finally, the four fans are reduced to the critical speed required to maintain stable adhesion, enabling the robot to firmly attach to the wall and enter wall-climbing mode. The robot can also complete a full cycle of transition, moving from the ground to the wall and back to the ground, as illustrated in the figure.

\subsection{Comprehensive Adsorption Capacity Level}

\begin{figure}[t]
	\begin{center}		
		\includegraphics[width=1.0\columnwidth]{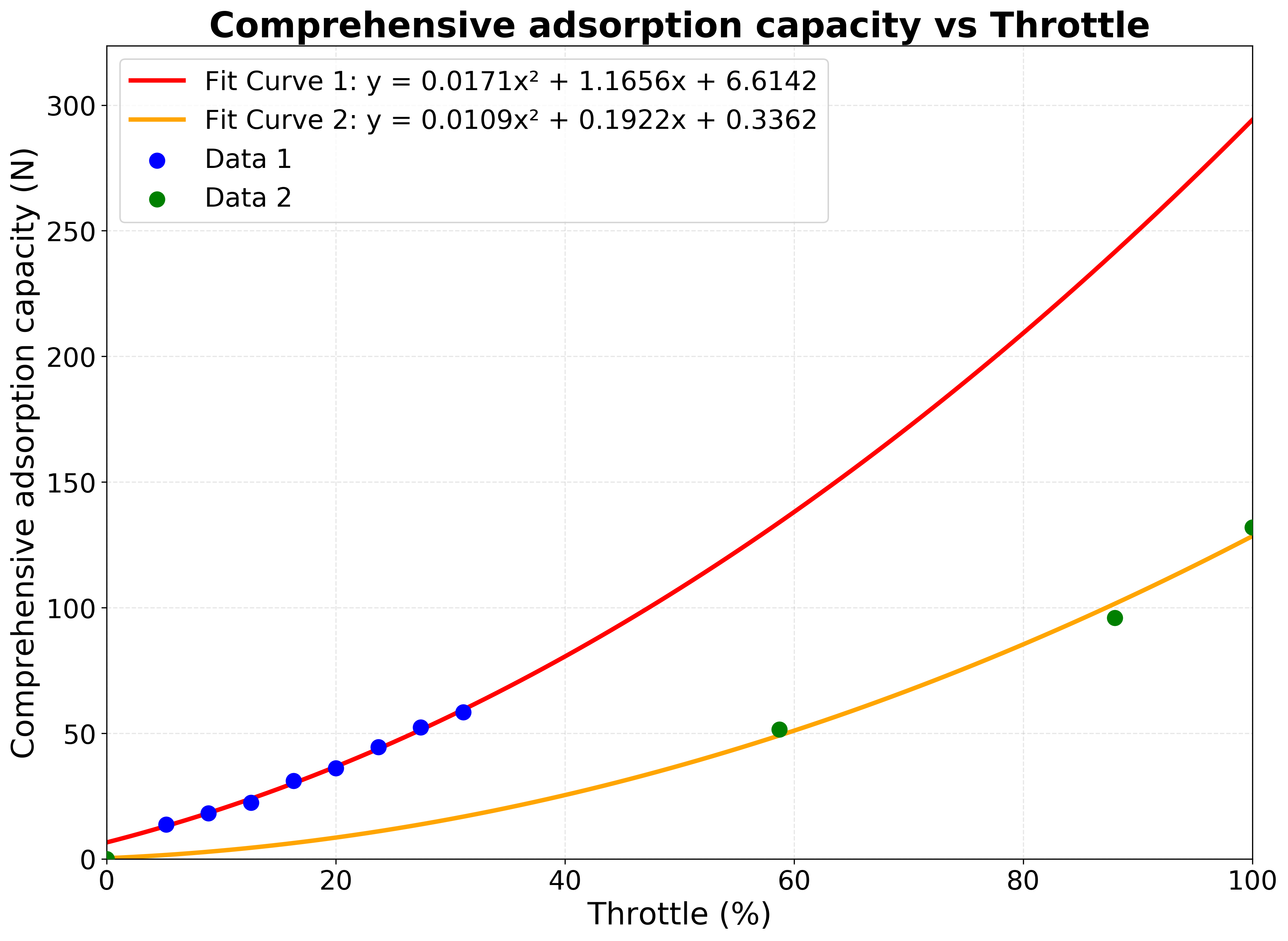}
	\end{center}
	\vspace{-0.5cm}
	\caption{
		\label{fig:Comprehensive Adsorption Capacity Level vs Throttle}
		Comprehensive Adsorption Capacity Level vs Throttle.
	}
	\vspace{-0.2cm}
\end{figure}

In the control modeling process, it is crucial to establish the quantitative relationship between the robot’s throttle input and the resulting overall adhesion force. To this end, a tensile test was performed under a constant-voltage power supply to evaluate the robot’s comprehensive adhesion capacity in vertical wall-climbing mode. The experimental results revealed a nonlinear correlation between throttle input and adhesion force, which can be well characterized by the quadratic fitting curve given in Eq.~\eqref{eq:Adsorption}. The corresponding dataset is illustrated in Fig.~\ref{fig:Comprehensive Adsorption Capacity Level vs Throttle}.

\begin{equation}
    \label{eq:Adsorption}
        \begin{aligned}
        f=0.017thr^2+1.166thr+6.614
        \end{aligned}
\end{equation}

The quadratic fitting curve 2 in Eq.~\eqref{eq:Adsorption} corresponds to the theoretical performance data provided by the ducted fan manufacturer. The deviation between curve 1 (experimental results) and curve 2 (theoretical data) represents the robot’s effective adhesion force. The maximum absolute deviation is observed at a throttle setting of 100\%, where the relative increase of the adhesion force reaches $165.68 \mathrm{N}$.

\subsection{Challenging Tasks}
PerchMobi\textsuperscript{3} successfully carried out a series of integrated experiments across ground, aerial, and wall domains. As illustrated in Fig.~\ref{fig:abstract}(a), the robot first traversed the ground toward a vertical wall. Upon the front wheels making contact with the wall, a flip maneuver was executed, after which the system adhered to the wall and transitioned into adhesion mode. Using its wheels, the robot then demonstrated upward, lateral, and downward locomotion along the wall surface. Once the wheels re-established contact with the ground, a reverse flip maneuver was performed, allowing the robot to detach and continue moving in ground mode. As shown in Fig.~\ref{fig:abstract}(b), the robot also demonstrated seamless ground-to-air transition: when encountering an obstacle during ground motion, it executed a vertical takeoff to enter aerial mode and successfully leaped over the obstacle.

One experimental scenario for PerchMobi\textsuperscript{3} is obstacle negotiation in complex environments: when encountering large gaps, the robot can activate flight mode to leap over them, while narrow wall seams can be traversed by leveraging flip and wall-climbing modes, as illustrated in Fig.~\ref{fig:introduction}.

\begin{figure}[t]
	\begin{center}
		\includegraphics[width=1.0\columnwidth]{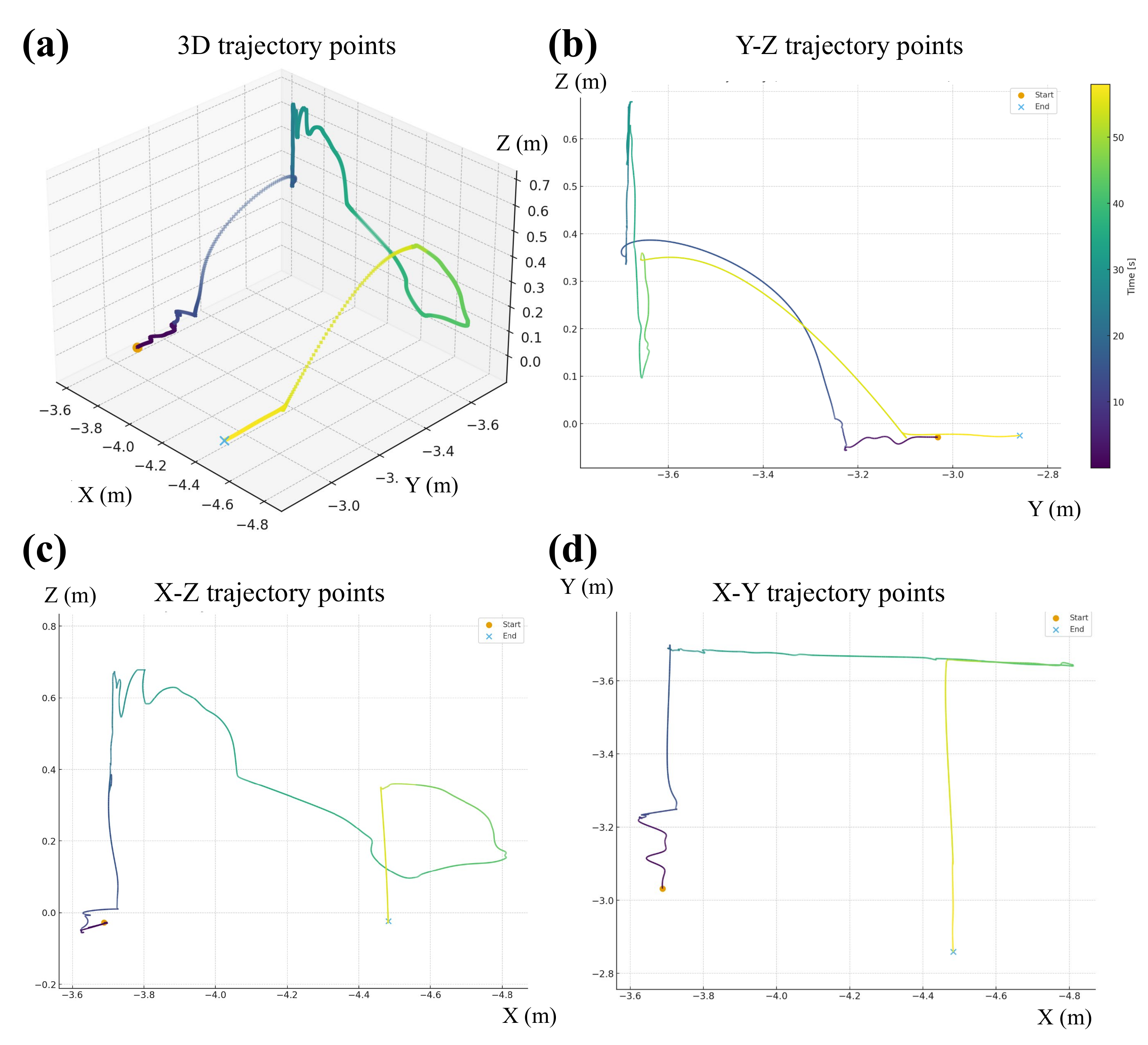}
	\end{center}
	%\vspace{-0.5cm}
	\caption{
		\label{fig:vicon}
		Trajectory points corresponding Fig.~\ref{fig:abstract}. (a) 3D trajectory points. (b) Y-Z trajectory points, on the right is a color bar representing time. (c) X-Z trajectory points, (d) X-Y trajectory points.
	}
	%\vspace{-0.2cm}
\end{figure}

\section{Conclusion}
This work introduced PerchMobi\textsuperscript{3}, a quad-fan, negative-pressure, air-ground-wall robot that demonstrates the feasibility of propulsion–adhesion power reuse. By repurposing four ducted fans to simultaneously provide aerial thrust and negative-pressure adhesion, and integrating them with four actively driven wheels under a dedicated modeling and control framework, the system achieves coordinated locomotion across ground, wall, and aerial domains with fan-assisted transitions. To the best of our knowledge, PerchMobi\textsuperscript{3} represents the first quad-fan prototype to realize functional power reuse for integrated multi-modal locomotion. A full-scale prototype was developed, and its feasibility was validated through comprehensive experiments in ground driving, payload-assisted wall climbing, aerial flight, and cross-mode transitions, confirming the effectiveness of the proposed design.  

Looking forward, future work will expand these feasibility demonstrations with systematic quantitative evaluations and comparative studies against existing wall-climbing and multi-modal robots. We also plan to enhance autonomy through onboard perception and planning for fully untethered operation, and to improve energy efficiency through refined fan–wheel coordination and optimized control strategies. In addition, adaptability to unstructured environments with diverse wall surfaces, complex transitions, and external disturbances will be further investigated. With these advancements, PerchMobi\textsuperscript{3} has the potential to evolve into a robust and versatile platform, establishing a new paradigm for energy-efficient and task-capable multi-modal robotic systems.

\bibliographystyle{ieeetr}
\bibliography{references}

\end{document}